\documentclass{article}
\usepackage[final,main]{neurips_2026}
\usepackage[utf8]{inputenc}
\usepackage[T1]{fontenc}
\usepackage{amsmath,amssymb}
\usepackage{graphicx,longtable,booktabs}
\usepackage{hyperref}
\usepackage{float}
\graphicspath{{assets/}}

\title{SENTRY: Deterministic, Intelligent Risk Assessment for IT Change Management\thanks{This work was completed at Royal Bank of Canada as part of the RBC Amplify program.}}
\author{
\textbf{Daniel Arulpragasam}\\Royal Bank of Canada\\\texttt{j.danielarul@gmail.com}
\and
\textbf{Christer Henrysson}\\Royal Bank of Canada\\\texttt{christer.jt.henrysson@gmail.com}
\and
\textbf{Ella Ly}\\Royal Bank of Canada\\\texttt{lyella15@gmail.com}
\and
\textbf{Deepika Anbalagan}\\Royal Bank of Canada\\\texttt{deepika.a5674@gmail.com}
\and
\textbf{Leo Feng}\\RBC Borealis\\\texttt{leo.feng@borealisai.com}
}
\date{}

\begin{document}
\maketitle

\begin{abstract}
Technology change management in large financial institutions depends on
risk assessments that are accurate, consistent, and auditable. In
practice, many institutions still rely on self-reported questionnaires.
Those questionnaires are subjective, easy to game, and poor at
separating routine changes from the ones that later trigger major
incidents. This paper presents SENTRY, a risk assessment platform that
replaces questionnaire-based scoring with a deterministic machine
learning pipeline built from gradient-boosted decision trees (XGBoost)
and hybrid retrieval-augmented generation (RAG). The system combines
structured operational metadata, application dependency graphs, and
historical incident records with a hybrid semantic and lexical search
over historical change requests. The retrieval step captures the risk
signal in unstructured change request text, then compresses that signal
into a single scalar feature before model inference. That design keeps
the model deterministic and preserves per-prediction explainability via
SHAP values. Evaluated on
enterprise-scale change data, SENTRY achieves a ROC AUC of 0.87 and
85\% overall accuracy, and it detects high-risk changes at roughly 3.25
times the rate of the existing process. We close by examining the
architectural trade-offs behind this design and what they imply for the
use of machine learning in regulated change management.

\end{abstract}

\section{Introduction}\label{introduction}

Technology changes such as software deployments, infrastructure
upgrades, configuration modifications, and security patches are routine
in financial institutions, but routine does not mean harmless. A misconfigured deployment can bring
down a trading platform, a poorly tested patch can take down a service, 
and an under-documented infrastructure change can ripple through
tightly coupled systems in ways that are hard to predict.

The institutional consequences are concrete. Financial institutions
process hundreds of thousands of change requests each year, and the
stakes of misclassification are high. Unfortunately, many major incidents (the kind
that trigger customer-facing outages, regulatory exposure, and material
financial loss) originate from changes that had been assessed as low
risk. The cost per incident compounds across regulatory fines,
operational losses, and reputational damage.

This failure is structural, change risk is
still assessed through a self-reported questionnaire. The process looks orderly, but it is fragile. It depends on the
accuracy of self-reported information, rewards users who know how to
minimize an apparent risk score, and yields labels that often say little
about why a change was judged low, medium, or high. It also misses the
context the requestor may not have: downstream application
dependencies, recent incident history, and the resemblance between the
proposed change and earlier changes that caused incidents.

Change approvers compensate for those gaps through manual investigation.
On average, a change request is reviewed in 10 minutes and each change request gets reviewed by approximately 4 approvers. During that
window, an approver may have to consult several disconnected enterprise
tools to verify dependency data, review incident history, or inspect the
evidence attached to the request. Those checks matter, but they are used
informally and rarely become part of the risk score itself. The result
is a labor-intensive process that becomes less consistent as volume
increases.

This paper presents SENTRY, a platform designed to replace
subjective questionnaire-based change risk assessment with an objective machine
learning pipeline that evaluates the operational risk of a proposed
change independently of the requestor's self-assessment. The central
contribution is architectural. We combine a supervised ML
classifier \citep{chen2016xgboost} operating 27 structured
features with a retrieval-augmented generation (RAG) layer
\citep{lewis2020rag} that captures the latent risk
signal in unstructured change request text. The RAG component performs
semantic and lexical search over a corpus of historical change requests
stored in a vector database, retrieves similar
historical changes together with their incident outcomes, and distills
the result into a single deterministic score that enters the
model as one feature. This allows the system to use information from
free-text fields such as descriptions, implementation plans, test plans,
and backout plans without giving up determinism, auditability, or
explainability via SHAP-style feature attributions
\citep{lundberg2017shap}.

The remainder of this paper is organized as follows. Section 2 reviews
related work and the limitations of existing approaches. Section 3
describes the system architecture. Section 4 details the XGBoost +
Section 5 presents evaluation results. Section 6 discusses
limitations, and Section 7 concludes.

\section{Related Work and
Background}\label{related-work-and-background}

\subsection{Traditional Change Risk
Assessment}\label{traditional-change-risk-assessment}

The IT Infrastructure Library (ITIL) framework, widely adopted in
enterprise IT service management (ITSM), prescribes a risk assessment step
within the change enablement process \citep{axelos2019itil}. In
practice, most organizations
implement this as a structured questionnaire completed by the change
requestor at submission time. The questions typically address impact
scope, testing completeness, backout readiness, and regulatory
sensitivity. Responses are scored and aggregated, often through simple
weighted sums, to produce a categorical risk rating.

The assessment remains subjective,
different reviewers can interpret the same questions differently, and
requestors who understand the scoring rules can shape answers to reduce
the apparent risk. In our dataset, the practical result is a heavy
concentration of ``low risk'' classifications, including for changes
that later cause incidents.

\subsection{Machine Learning
Approaches}\label{machine-learning-approaches}

The application of machine learning to IT operations (AIOps) has gained
traction in recent years, particularly for incident prediction, anomaly
detection, and root cause analysis \citep{dang2019aiops}. However, the
application of ML to change risk assessment specifically remains
underexplored in academic literature. Within the institution where
SENTRY was developed, several candidate approaches were systematically
evaluated before arriving at the current design.

The most natural starting point was large language model scoring. An LLM
can read unstructured change request text directly, reason across
multiple fields simultaneously. For a domain where much of the signal lives
in free-text implementation plans and backout descriptions, that is a
genuinely attractive property. However,
LLMs are non-deterministic, identical inputs can yield different outputs
across invocations, and the reasoning pathway from input to output is
not recoverable from the model. In a regulated financial institution,
where risk scores must be reproducible and traceable, both of those
properties are disqualifying. An auditor cannot accept a risk
classification whose value would change if the system were rerun an
hour later.

Attention turned next to more expressive supervised models. Deep neural
networks can represent highly non-linear feature interactions and are deterministic. That resolves the
reproducibility objection raised against LLMs. The explainability
objection remains, however. A multi-layer network does not offer a
natural decomposition of its prediction into per-feature contributions,
and post-hoc attribution methods for neural networks introduce their own
approximation errors. In a setting where change approvers need to
understand \emph{why} a change was flagged before acting on the score,
a model that is deterministic but opaque is not sufficient.

Tree-based ensemble methods address both concerns simultaneously. By
construction, a decision tree partitions the input space along feature
boundaries, so its predictions are fully deterministic and the path from
input to output can be traced exactly. Ensemble methods such as gradient
boosted trees extend this to capture the kinds of non-linear
interactions that a single shallow tree would miss. Among
tree-based alternatives, gradient boosting in particular has been shown
to perform well on structured, tabular data of the kind that dominates
change management records \citep{chen2016xgboost}.

XGBoost was selected as the final model on two grounds. First, it is
fully deterministic: given a fixed dataset, fixed hyperparameters, and a
fixed random seed, every training run and every inference call produces
identical output. Second, it exposes a native \texttt{pred\_contribs}
API that returns SHAP-style per-feature contribution values at inference
time \citep{lundberg2017shap}, making per-prediction explainability a
first-class output rather than an afterthought. LightGBM
\citep{ke2017lightgbm}, a close gradient-boosted alternative, was also
evaluated on the same dataset but returned lower precision, recall, and
F1 across multiple configurations. XGBoost was therefore the candidate
that satisfied all three constraints simultaneously: determinism,
explainability, and empirical performance.

\subsection{Retrieval-Augmented
Generation}\label{retrieval-augmented-generation}

\citet{lewis2020rag} define retrieval-augmented generation (RAG) as a
paradigm in which a retrieval system fetches relevant documents from a
knowledge base, and the
retrieved content is used to augment the input to a downstream model.
RAG is usually discussed in the context of grounding LLM outputs in
factual content, but the retrieval step is useful on its own because it
can identify semantically similar historical records.

In our architecture, we adapt the RAG concept: we retrieve historically
similar change requests using hybrid search (combining dense vector
embeddings with sparse lexical matching), but rather than feeding the
retrieved records into an LLM, we compute a deterministic risk score
from the retrieval results and pass it as a scalar feature to the
XGBoost model. This approach captures the semantic richness of RAG while
maintaining the determinism and explainability of a traditional ML
pipeline. To our knowledge, that combination has not previously been
applied to IT change risk assessment.

\section{System Design}\label{system-design}

\subsection{Architecture Overview}\label{architecture-overview}

SENTRY is organized around four components: a web frontend for user
interaction, a backend API layer for orchestration, a trained machine
learning model for risk classification, and separate data stores for
structured ITSM records and vector-based similarity search.

For the purposes of this paper, we will only observe the backend, specifically the ML model.

The backend is an API service that coordinates the analysis. For each
change request number, it queries several enterprise data sources in
parallel: a relational database replica of the institution's ITSM
platform, an application portfolio management service for application
metadata and dependency mapping and a vector database for similarity
search.

\begin{figure}[t]
\centering
\includegraphics[width=0.8\linewidth]{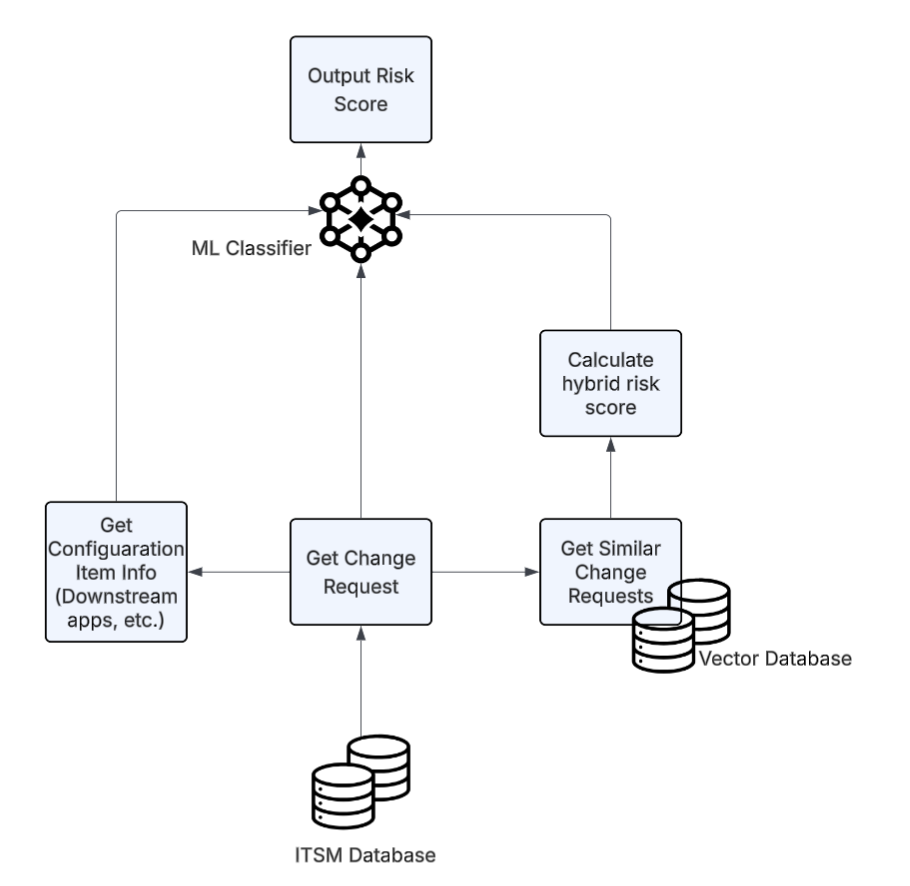}
\caption{High-level architecture of the SENTRY risk assessment platform.}
\label{fig:architecture}
\end{figure}

\subsection{Data Sources}\label{data-sources}

SENTRY draws from four external data sources:

\begin{enumerate}
\def\labelenumi{\arabic{enumi}.}
\item
  \textbf{ITSM database}: Contains change request records, configuration
  items, application codes, and incident data across software and
  infrastructure assets.
\item
  \textbf{Application portfolio management service}: Provides application
  metadata including business criticality, crown jewel designation,
  SOX-critical flags, data classification, recovery time objectives (RTO),
  user counts, and the downstream dependency graph used for blast radius
  analysis.
\item
  \textbf{Vector database}: Stores dense embeddings of historical change
  request text fields alongside a precomputed lexical index. This database
  powers the hybrid RAG pipeline described in Section~\ref{methodology-xgboost-with-hybrid-rag}.
\end{enumerate}

\section{Methodology: XGBoost with Hybrid
RAG}\label{methodology-xgboost-with-hybrid-rag}

This section describes the core of SENTRY's risk engine. Unstructured
text fields in change requests, including descriptions, implementation
plans, test plans, and backout plans, contain real risk signal. Raw
text embeddings, however, did not behave well as direct tabular
features and made the model harder to explain. We therefore use hybrid
search to retrieve historically similar changes, convert the retrieval
results into a risk-weighted score, and pass that score into XGBoost as
a single deterministic feature.

\subsection{Feature Set}\label{feature-set}

The XGBoost model operates on 28 features, all normalized to the {[}0,
1{]} range with capped maximums. The features are drawn from three
structured data sources plus the hybrid search output. They fall into
five groups that mirror how change approvers already reason about
operational exposure: application context, blast radius, incident
history, similarity to earlier changes, and change type.

\subsection{Hybrid RAG Pipeline}\label{hybrid-rag-pipeline}

The hybrid RAG pipeline transforms unstructured change request text into
a deterministic risk feature through three stages: embedding generation,
hybrid retrieval, and risk score computation.

\paragraph{Embedding Generation}\label{embedding-generation}

For each change request, five text fields (short description,
description, implementation plan, test plan, and backout plan) are
concatenated and embedded using a large text embedding model, producing
a high-dimensional dense vector. These embeddings are stored in a vector
database alongside a precomputed lexical index built from the same text
fields.

The embedding store uses a composite key structure that allows a single
change request to appear multiple times when it caused multiple
incidents. Change requests with no associated incident are stored with a
null incident reference. This denormalized structure lets the retrieval
query return incident metadata directly, without an additional join.

\paragraph{Hybrid Search}\label{hybrid-search}

When scoring a new change request, the system performs a hybrid search
that combines two retrieval strategies through Reciprocal Rank Fusion
(RRF) \citep{cormack2009rrf}.

The \textbf{semantic search} component retrieves the top-$k$ nearest
neighbors by cosine similarity over the dense embedding space. This
captures changes that are conceptually similar even without shared
terminology. A ``database schema migration'' and a ``table
restructuring,'' for instance, score highly against each other despite
minimal keyword overlap.

The \textbf{lexical search} component scores candidates using a
full-text ranking function against a precomputed lexical index. This
captures changes that share specific technical terms, application
names, or procedural language that the semantic model might
under-weight.

The two ranked lists are merged, and each result's fused score is
computed as:

\[
\mathrm{rrf\_score} = \alpha \times \frac{1}{60 + \mathrm{semantic\_rank}}
+ \beta \times \frac{1}{60 + \mathrm{lexical\_rank}}
\]

The constant 60 is the standard RRF smoothing parameter that prevents
top-ranked results from dominating. The $\alpha/\beta$ weighting gives lexical
search slightly higher influence. This reflects an empirical observation
that keyword matches tend to surface more
operationally relevant results in the ITSM domain, where specific
application names and technical terms carry strong risk signals. If a
result appears in only one retrieval list, the missing rank contributes
zero to the fused score.

To prevent data leakage during model training, the query excludes the
current change request by number and filters to change requests opened
before the query change request's creation date.

\paragraph{Risk Score Computation}\label{risk-score-computation}

Once the top-k most similar change requests are retrieved (default
k=100), the hybrid search score is computed as a weighted sum over all
incident-bearing neighbors:

\[
\mathrm{hybrid\_search\_score} = \sum (\mathrm{rrf\_score} \times
\mathrm{priority\_points})
\]

where the priority points use a scale tuned for the model where higher priority
incidents equal more points. Change requests with no associated
incident contribute zero to the score.

The intuition is straightforward: if a proposed change is textually
similar to many historical changes that caused high-severity incidents,
the hybrid search score will be high; if its nearest neighbors are
benign, the score will be low. This captures a risk dimension that
structured metadata alone cannot capture: the nature of the change
itself, as described by the humans who planned it.

\begin{figure}[t]
\centering
\includegraphics[width=0.65\linewidth]{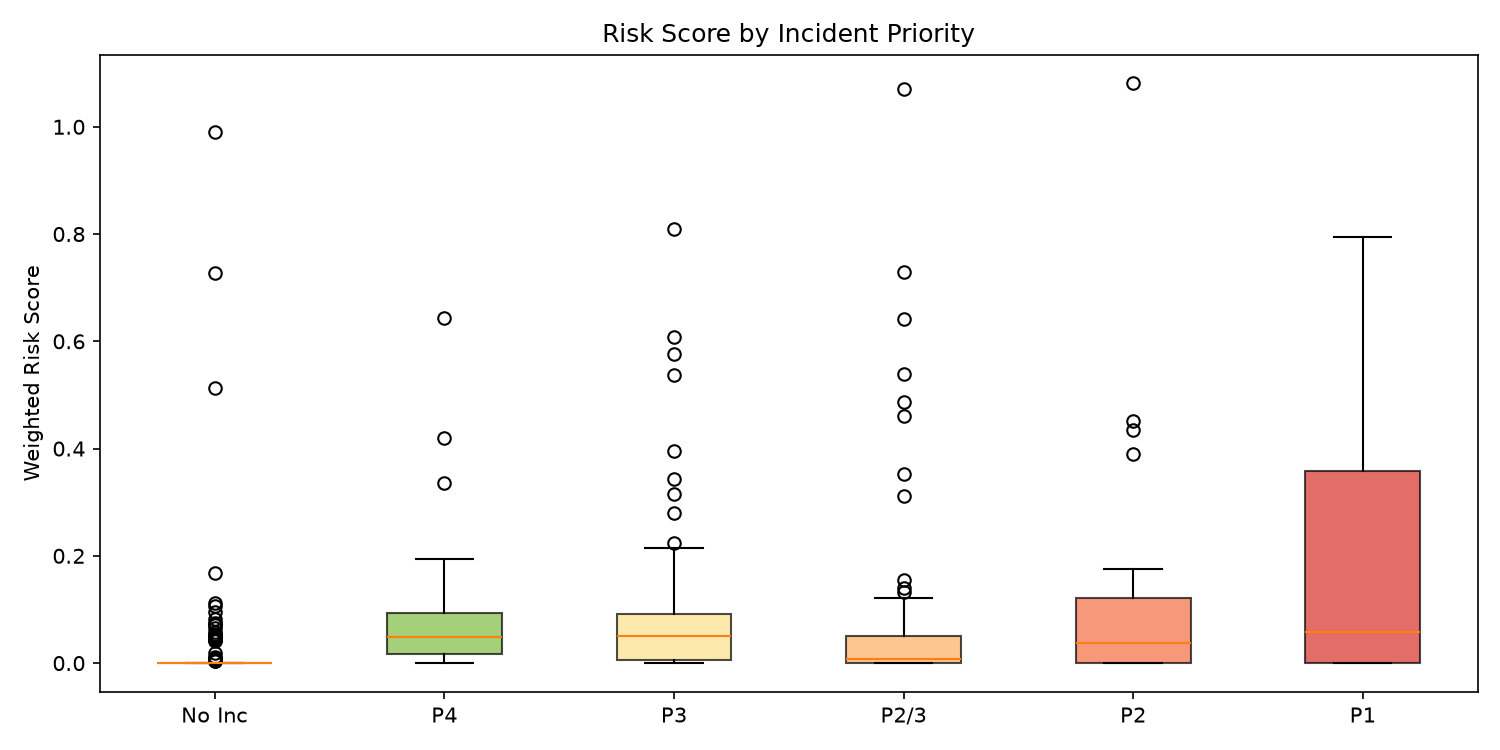}
\caption{Hybrid risk score based on changes and the respective incident they caused. $\alpha=0.4$ $\beta=0.6$ .}
\label{fig:hybrid_rag_effectiveness}
\end{figure}
\paragraph{Why This Architecture}\label{why-this-architecture}

An earlier iteration of the system attempted to incorporate PCA
embedding features directly as features in the XGBoost
model. This failed for two practical reasons. The embedding
dimensions dominated the feature space, drowning out the other structured
features and reducing overall model accuracy. The resulting feature
importance rankings were also dominated by opaque
\texttt{embedding\_1,\ embedding\_2,\ ...,\ embedding\_12} entries that
offered no interpretive value to change approvers trying to understand
why a change was flagged.

Our hybrid approach keeps the useful part of retrieval and discards the
unstable part. Figure~\ref{fig:hybrid_rag_effectiveness} was derived from 
calculating the risk score across 200 changes that caused no incidents and 
200 changes that caused incidents. Changes with caused incidents have a higher weighted 
risk score, indicating that unstructured text contains a meaningful signal. 

Because the final output is a deterministic
scalar that feeds into a fully interpretable model, the approver
sees ``\texttt{hybrid\_search\_score} contributed +0.048 to the risk
prediction'' alongside contributions from structured features such as
``\texttt{crown\_jewel} contributed +0.05.'' That level of
explainability is difficult to obtain from either raw embeddings or
LLM-generated assessments.

\subsection{Model Training}\label{model-training}

Training is orchestrated through an automated pipeline with
hyperparameter optimization:

\textbf{Dataset preparation.} The training dataset is constructed from
labeled change request records exported from the institution's
ITSM database. Positive examples (label 1, ``High/Medium risk'')
consist of all change requests that caused major incidents,
approximately 183 records spanning January 2021 to June 2026. Negative
examples (label 0, ``Low risk'') are sampled from change requests
associated with minor incidents and change requests with no incident at
all, drawn proportionally to maintain an approximately 80/20 class
ratio. All label-1 records are included without downsampling; label-0
records are capped per source category to prevent any single
subpopulation from dominating the training signal. A stratified
train/test split is applied.

\textbf{Feature extraction.} For each change request in the training
set, the full 28-feature vector is computed by querying live data
sources. A temporal filter on the hybrid search ensures that only
change requests opened \emph{before} the target CR are considered,
preventing data leakage.

\textbf{Hyperparameter optimization.} We use Optuna
\citep{akiba2019optuna} with a Tree-structured Parzen Estimator (TPE)
sampler to search over 60 trial configurations. The search space
includes \texttt{n\_estimators}
(50--400), \texttt{max\_depth} (2--8), \texttt{learning\_rate}
(0.01--0.3, log scale), \texttt{min\_child\_weight} (1--20),
\texttt{subsample} (0.6--1.0), \texttt{colsample\_bytree} (0.5--1.0),
\texttt{gamma} (0.0--5.0), \texttt{reg\_alpha} (1e-8--10.0, log scale),
and \texttt{reg\_lambda} (1e-8--10.0, log scale). Each trial is
evaluated using 5-fold stratified cross-validation.

\textbf{Objective function.} The optimization objective maximizes F1
score (positive class) with a secondary penalty for precision-recall
imbalance:
\texttt{objective\ =\ F1\ -\ (\textbar{}precision\ -\ recall\textbar{}\ *\ 0.1)}.
This encourages the optimizer to find configurations that balance
catching high-risk changes (recall) with avoiding excessive false alarms
(precision), a tension that is particularly relevant in change
management, where flooding the Change Advisory Board with false
positives erodes trust, while missing genuinely high-risk changes has
severe operational consequences.

\textbf{Threshold calibration.} The model outputs a continuous
probability P(High/Med) for each change request. Two thresholds map this
probability to a three-tier classification: High risk ($P \ge 0.90$,
fixed), Medium risk ($P \ge \text{medium threshold}$, auto-selected
during training), and Low risk ($P < \text{medium threshold}$). The
medium threshold is
optimized during training by evaluating candidates in the range
0.30--0.75, selecting the value that minimizes the absolute gap between
precision and recall, with F1 as a tiebreaker.

\textbf{Class imbalance handling.} The severe class imbalance inherent
in the problem (the vast majority of changes do not cause incidents) is
addressed through \texttt{scale\_pos\_weight}, which is computed
automatically as the ratio of negative to positive examples in the
training set.

\section{Evaluation}\label{evaluation}

\subsection{Model Performance}\label{model-performance}

The XGBoost model was evaluated on a held-out test set of 203 change
requests using standard classification metrics:

\begin{table}[h]
\centering
\caption{Confusion matrix on the held-out test set ($n = 203$).}
\label{tab:confusion}
\begin{tabular}{lcc}
\toprule
 & \textbf{Predicted Low} & \textbf{Predicted High/Med} \\
\midrule
\textbf{Actual Low} & 150 & 18 \\
\textbf{Actual High/Med} & 13 & 22 \\
\bottomrule
\end{tabular}
\end{table} 

\begin{table}[h]
\centering
\caption{SENTRY vs.\ the questionnaire-based baseline on high-risk
detection.}
\label{tab:comparison}
\begin{tabular}{lcccc}
\toprule
\textbf{Method} & \textbf{\% of Incident Causing Changes Caught} \\
\midrule
Questionnaire-based & 20\% \\
SENTRY (XGBoost + Hybrid RAG) & 63\% \\
\bottomrule
\end{tabular}
\end{table}

\begin{figure}[t]
\centering
\includegraphics[width=0.65\linewidth]{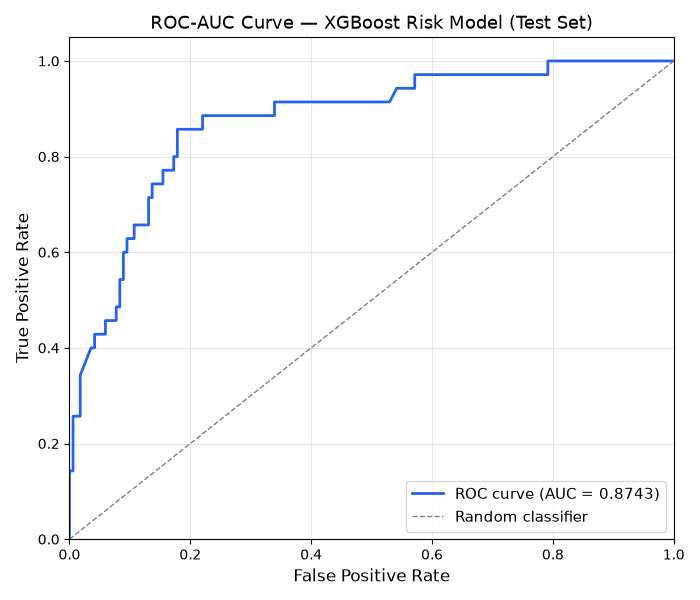}
\caption{ROC curve on the held-out test set (AUC\,=\,0.87).}
\label{fig:roc}
\end{figure}

\begin{table}[h]
\centering
\caption{SENTRY per-class precision, recall, and F1 on the held-out test set.}
\label{tab:perclass}
\begin{tabular}{lcccc}
\toprule
\textbf{Class} & \textbf{Precision} & \textbf{Recall} & \textbf{F1} & \textbf{Support} \\
\midrule
Low risk & 0.92 & 0.89 & 0.91 & 168 \\
High/Medium risk & 0.55 & 0.63 & 0.59 & 35 \\
\bottomrule
\end{tabular}
\end{table}

The most important result is recall on the positive class. The model
correctly identifies 63\% of changes that go on to cause major
incidents as medium or high risk, a 3.25$\times$ improvement over the
existing questionnaire-based process, which correctly identifies
approximately 20\% of such changes.

That improvement comes with a real trade-off. Precision on the positive
class is 0.55, which means some changes flagged for additional scrutiny
will not go on to cause major incidents. In this setting, that is an
acceptable compromise. Financial institutions cannot afford to miss
high-risk changes, but they also cannot flood the Change Advisory Board
with so many false positives that reviewers stop trusting the tool. The
F1-optimized training objective with a precision-recall balance penalty
was designed around that operational tension. \\ \\

\subsection{Feature Importance}\label{feature-importance}

Gain-based feature importance rankings confirm that the hybrid search
score is a meaningful risk predictor. Table~\ref{tab:feature_importance}
shows the top-8 features by gain from the best-performing Optuna trial.
\texttt{hybrid\_search\_score} ranks fourth, sitting alongside three
structured metadata features in the top five.

\begin{table}[h]
\centering
\caption{Top-8 features by gain-based importance (best Optuna trial).
The hybrid search score ranks 4th, confirming that unstructured text
adds signal beyond structured metadata alone.}
\label{tab:feature_importance}
\begin{tabular}{clc}
\toprule
\textbf{Rank} & \textbf{Feature} & \textbf{Gain (\%)} \\
\midrule
1 & \texttt{type\_standard} & 15.3 \\
2 & \texttt{crown\_jewel} & 5.4 \\
3 & \texttt{inherent\_risk\_score} & 5.0 \\
4 & \texttt{hybrid\_search\_score} & 4.9 \\
5 & \texttt{num\_deps} & 4.8 \\
6 & \texttt{data\_classification} & 4.7 \\
7 & \texttt{p2\_2yr} & 4.5 \\
8 & \texttt{biz\_lines} & 4.3 \\
\bottomrule
\end{tabular}
\end{table}

\section{Discussion and Limitations}\label{discussion-and-limitations}

Several limitations of the current system warrant discussion.

\textbf{Limited positive examples.} The training dataset contains
approximately 183 label-1 examples (change requests that caused P1/P2
incidents). While this reflects the genuine scarcity of major incidents
relative to the volume of changes, it constrains the model's ability to
learn fine-grained distinctions among high-risk changes. Expanding the
training window or synthesizing additional positive examples through
controlled augmentation are potential avenues for improvement.

\textbf{Static training data.} The model is trained on a fixed snapshot
of historical data. As the institution's technology landscape evolves,
with new applications, shifting dependency structures, and emerging
change patterns, the model's predictive accuracy will degrade unless it is
periodically retrained. An automated retraining pipeline is planned but
not yet implemented.

\textbf{Temporal scope of embeddings.} The vector database requires
manual updates to incorporate new change requests into the hybrid search
corpus. Until this process is automated, the RAG pipeline operates on a
static snapshot that may not reflect the most recent change activity.

\textbf{Generalizability.} While the methodology is domain-general, the
specific feature definitions, normalization ranges, and threshold values
are calibrated to the operational environment of a single large
financial institution. Deployment in a different organization would
require re-extraction of features, re-calibration of normalization
parameters, and retraining of the model on local data.

\section{Conclusion}\label{conclusion}

This paper presented SENTRY, a deterministic risk assessment platform
for IT change management that combines XGBoost with hybrid
retrieval-augmented generation. The central challenge was not simply to
improve prediction. It was to use the risk-relevant information embedded
in unstructured change request text without sacrificing the determinism,
auditability, and explainability that regulated environments require.

Our approach uses hybrid semantic and lexical search to retrieve
historically similar changes, computes a risk-weighted score from those
retrieval results, and feeds that score into an XGBoost classifier as a
single feature. This resolves the tension between expressiveness and
determinism that has limited prior approaches. The resulting model
operates on 28 features spanning application metadata, blast radius,
incident history, and the hybrid search score, producing a three-tier
risk classification with per-prediction SHAP explanations.

Evaluated on enterprise-scale change data from a major financial
institution, the system achieves a ROC AUC of 0.87 and detects high-risk
changes at 3.25 times the rate of the existing questionnaire-based
process. 

As IT change volumes continue to grow and the complexity of enterprise
technology ecosystems deepens, the need for objective, evidence-driven
risk assessment will only intensify. SENTRY is best understood as a
decision-support system for human reviewers. It does not automate
approval decisions. It gives those reviewers better information, faster,
and with more transparency than the process they use today.

\newpage
\bibliographystyle{abbrvnat}
\bibliography{SENTRY}

\begin{thebibliography}{8}
\providecommand{\natexlab}[1]{#1}
\providecommand{\url}[1]{\texttt{#1}}
\expandafter\ifx\csname urlstyle\endcsname\relax
  \providecommand{\doi}[1]{doi: #1}\else
  \providecommand{\doi}{doi: \begingroup \urlstyle{rm}\Url}\fi

\bibitem[Akiba et~al.(2019)Akiba, Sano, Yanase, Ohta, and
  Koyama]{akiba2019optuna}
T.~Akiba, S.~Sano, T.~Yanase, T.~Ohta, and M.~Koyama.
\newblock Optuna: A next-generation hyperparameter optimization framework.
\newblock In \emph{Proceedings of the 25th ACM SIGKDD International Conference
  on Knowledge Discovery and Data Mining}, pages 2623--2631, 2019.

\bibitem[{AXELOS}(2019)]{axelos2019itil}
{AXELOS}.
\newblock \emph{ITIL Foundation: ITIL 4 Edition}.
\newblock TSO, 2019.

\bibitem[Chen and Guestrin(2016)]{chen2016xgboost}
T.~Chen and C.~Guestrin.
\newblock Xgboost: A scalable tree boosting system.
\newblock In \emph{Proceedings of the 22nd ACM SIGKDD International Conference
  on Knowledge Discovery and Data Mining}, pages 785--794, 2016.

\bibitem[Cormack et~al.(2009)Cormack, Clarke, and Buettcher]{cormack2009rrf}
G.~V. Cormack, C.~L.~A. Clarke, and S.~Buettcher.
\newblock Reciprocal rank fusion outperforms condorcet and individual rank
  learning methods.
\newblock In \emph{Proceedings of the 32nd International ACM SIGIR Conference
  on Research and Development in Information Retrieval}, pages 758--759, 2009.

\bibitem[Dang et~al.(2019)Dang, Lin, and Huang]{dang2019aiops}
Y.~Dang, Q.~Lin, and P.~Huang.
\newblock {AIOps}: Real-world challenges and research innovations.
\newblock In \emph{Proceedings of the 41st International Conference on Software
  Engineering: Companion Proceedings}, pages 4--5, 2019.

\bibitem[Ke et~al.(2017)Ke, Meng, Finley, Wang, Chen, Ma, Ye, and
  Liu]{ke2017lightgbm}
G.~Ke, Q.~Meng, T.~Finley, T.~Wang, W.~Chen, W.~Ma, Q.~Ye, and T.-Y. Liu.
\newblock {LightGBM}: A highly efficient gradient boosting decision tree.
\newblock In \emph{Advances in Neural Information Processing Systems 30}, pages
  3146--3154, 2017.

\bibitem[Lewis et~al.(2020)Lewis, Perez, Piktus, Petroni, Karpukhin, Goyal,
  K{"u}ttler, Lewis, Yih, Rockt{"a}schel, Riedel, and Kiela]{lewis2020rag}
P.~Lewis, E.~Perez, A.~Piktus, F.~Petroni, V.~Karpukhin, N.~Goyal,
  H.~K{"u}ttler, M.~Lewis, W.~Yih, T.~Rockt{"a}schel, S.~Riedel, and D.~Kiela.
\newblock Retrieval-augmented generation for knowledge-intensive {NLP} tasks.
\newblock In \emph{Advances in Neural Information Processing Systems 33}, pages
  9459--9474, 2020.

\bibitem[Lundberg and Lee(2017)]{lundberg2017shap}
S.~M. Lundberg and S.~Lee.
\newblock A unified approach to interpreting model predictions.
\newblock In \emph{Advances in Neural Information Processing Systems 30}, pages
  4765--4774, 2017.

\end{thebibliography}

\appendix

\end{document}